# A Comparison of Rule-Based and Deep Learning Models for Patient Phenotyping

**Sebastian Gehrmann[0,1*], Franck Dernoncourt[0,2], Yeran Li[0,3], Eric T Carlson[0,4], Joy T Wu[0,5], Jonathan Welt[0,6], John Foote Jr.[0,7], Edward Moseley[0,8], David W Grant[0,9], Patrick D Tyler[0,5], Leo Anthony Celi[0,2]**

[0]**MIT Critical Data, Laboratory for Computational Physiology**

[1]**Harvard John A. Paulson School of Engineering and Applied Sciences**

[2]**Massachusetts Institute of Technology**

[3]**Harvard T.H. Chan School of Public Health**

[4]**Philips Research North America**

[5]**Beth Israel Deaconess Medical Center**

[6]**Massachusetts General Hospital**

[7]**Tufts University School of Medicine**

[8]**University of Massachusetts**

[9]**Washington University School of Medicine**

**Abstract**

**Objective**: We investigate whether deep learning techniques for natural language processing (NLP) can be used efficiently for patient phenotyping. Patient phenotyping is a classification task for determining whether a patient has a medical condition, and is a crucial part of secondary analysis of healthcare data. We assess the performance of deep learning algorithms and compare them with classical NLP approaches.

**Materials and Methods**: We compare convolutional neural networks (CNNs), n-gram models, and approaches based on cTAKES that extract pre-defined medical concepts from clinical notes and use them to predict patient phenotypes. The performance is tested on 10 different phenotyping tasks using 1,610 discharge summaries extracted from the MIMIC-III database.

**Results**: CNNs outperform other phenotyping algorithms in all 10 tasks. The average F1-score of our model is 76 (PPV of 83, and sensitivity of 71) with our model having an F1-score up to 37 points higher than alternative approaches. We additionally assess the interpretability of our model by presenting a method that extracts the most salient phrases for a particular prediction.

**Conclusion**: We show that NLP methods based on deep learning improve the performance of patient phenotyping. Our CNN-based algorithm automatically learns the phrases associated with each patient phenotype. As such, it reduces the annotation complexity for clinical domain experts, who are normally required to develop task-specific annotation rules and identify relevant phrases. Our method performs well in terms of both performance and interpretability, which indicates that deep learning is an effective approach to patient phenotyping based on clinicians' notes.

[*] gehrmann@seas.harvard.edu

# INTRODUCTION

The secondary analysis of data from electronic health records (EHRs) is crucial to better understand the heterogeneity of treatment effects and to individualize patient care. With the growing adoption rate of EHRs,[1] researchers gain access to rich data sets, such as the Medical Information Mart for Intensive Care or MIMIC database[2,3] and the Informatics for Integrating Biology and the Bedside (i2b2) datamarts,[4–9] which can be explored in numerous ways.[10] EHR data comprise both structured data such as International Classification of Diseases (ICD) codes, laboratory results and medications, and unstructured data such as clinician progress notes. While structured data do not require complex processing prior to performing statistical tests and conducting machine learning tasks, the majority of recorded data exist in unstructured form.[11] Applying natural language processing (NLP) on the unstructured data in conjunction with analyzing the structured data can lead to a better understanding of health and diseases,[12] and a more accurate phenotyping of patients to compare tests and treatments.[13–15] Patient phenotyping is a classification task to determine whether a patient has a medical condition, or pinpointing patients who are at risk for developing one. Further, intelligent applications for patient phenotyping can support clinicians by reducing the time they spend on chart reviews, which takes up a significant fraction of their daily workflow.[16,17]

A popular approach to patient phenotyping using NLP is based on extracting relevant medical phrases from texts and using them as input to build a predictive model.[18] The dictionary of relevant phrases is task-specific and its development requires significant effort and a deep understanding of the task from domain experts.[19] A different and more involved approach is to develop a rule-based algorithm for each condition.[20] Due to the tedious and laborious task required of clinicians to build a generalizable model for patient phenotyping, models for automated classification using NLP are rarely developed outside of the research area. However, recent developments in deep learning may provide an opportunity to build a generalizable phenotyping model with a less intense domain expert involvement. Applications of deep learning in healthcare have shown promising results; examples include mortality prediction,[21] patient note de-identification,[22] skin cancer detection,[23] and diabetic retinopathy detection.[24]

A drawback to deep learning models is their lack of interpretability. Interpretability means that one can understand how the features of the model arrive at the predictions. Since results directly impact health, clinicians have come to expect healthcare applications to use interpretable models.[25] Moreover, the European Union is considering regulations that require algorithms to be interpretable.[26] While much work has been done to understand deep learning NLP models and make a trained deep learning NLP model interpretable,[27–29] they rely on complex interactions between all inputs and are thus inherently less interpretable than an NLP model that uses predefined phrase dictionaries.

In this work, we investigate the application of convolutional neural networks (CNNs)[30] to text-based patient phenotyping. CNNs learn to identify phrases in text that lead to a positive or negative classification, similar to the phrase dictionary approach, and they outperform traditional approaches to classification problems in other domains.[31–33] We compare CNNs to the traditional rule-based entity extraction systems using the Mayo clinical Text Analysis and Knowledge Extraction System (cTAKES),[34] and other NLP methods such as logistic regression models using n-gram features. We compare the performance for a total of 10 different phenotypes and show that CNNs outperform both extraction-based and n-gram-based methods. Finally, we evaluate the interpretability of the model by assessing the learned phrases that are associated with each phenotype and compare them to the phrase dictionaries developed by clinicians.

# BACKGROUND

Accurate patient phenotyping is required for secondary analysis of EHRs to correctly identify the patient cohort under investigation and to better identify the clinical context.[35] Studies employing a manual chart review process for patient phenotyping are naturally limited to a small number of preselected patients. Therefore, NLP is necessary to identify information that is contained in text but may be inconsistently captured with accuracy in the structured data, such as recurrence in cancer,[36,37] whether a patient smokes,[4] classification within the autism spectrum,[38] or drug treatment patterns.[39] However, unstructured data in EHRs, such as progress notes or discharge summaries, is typically not amenable to simple text searches because of spelling mistakes, and the use of ambiguous terms.[40] To help address these issues, researchers utilize dictionaries and ontologies for medical terminologies such as UMLS[41] and SNOMed.[42]

Examples of the systems that employ such databases are the KnowledgeMap Concept Identifier (KMCI),[44] MetaMap,[45] and the cTAKES. These three identify words or phrases within a text and provide the medical concepts they are linked to.[34,46] They significantly reduce the work required from data scientists, who previously had to develop task-specific extractors.[47] Extracted entities are filtered to only include concepts related to the patient phenotype under investigation and either used as features for a model that predicts whether the patient fits the phenotype, or as input for rule-based algorithms.[18,38,48] Liao et al.[12] describe the process of extraction, rule-generation and prediction as the general approach to patient phenotyping using the cTAKES,[13,49–51] and test this approach on various data sets.[52] The role of clinicians in this task is to develop a task-specific dictionary of phrases that are relevant to a patient phenotype.

Carrell et al.[36] developed two separate rule-based phenotyping algorithms, one for pathology documents and one for clinical documents, which they combined in order to identify recurrence of breast cancer. While they find that this modular approach identifies over 90% of recurrence, they note that the cost and time required to develop an NLP algorithm limits its applicability to large or repeated tasks. Moreover, while a usable system would offset the development costs, it does not address the problem that a different specialized NLP system would have to be developed for every task in a hospital.

Halpern et al.[15] address the heavy workload for clinicians and describe a semi-supervised approach to this problem that uses the *Anchor and Learn Framework*.[53] In this scheme, the clinicians only need to define a few anchors, which are phrases that identify concepts with a very high positive predictive value (PPV). They train a supervised model that uses a combination of structured data and a bag-of-words of the notes to predict whether such anchor exists in a note. They showed that their method drastically reduces the required effort for clinicians while yielding equivalent results. Our supervised approach aims to reduce complexity for clinicians while achieving both a high PPV and sensitivity to correctly capture the whole patient cohort. Furthermore, we develop our algorithm to create a phrase dictionary to use for patient phenotyping, and compare it to cTAKES-based models.

# METHODS

## Concept-Extraction-Based Models

For our baseline models, we use cTAKES to extract concepts from each note. In cTAKES, sentences and phases are split into tokens (individual words). Then, tokens with variations (e.g. plural) are normalized to their base form. The normalized tokens are tagged for their part-of-speech (e.g. noun, verb), and a shallow parse tree is constructed to represent the grammatical structure of a sentence. Finally, a named-entity recognition (NER) algorithm uses this information to detect named entities that exist as concept unique identifiers (CUIs) in UMLS.[41]

While traditionally the rules were mostly fully hand-crafted, modern methods use relevant concepts in a note as input to a machine learning algorithm to directly learn to predict a phenotype.[54,55] Therefore, we specify two different approaches to using the cTAKES output. The first approach uses the complete list of extracted CUIs as input to further processing steps. In the second approach, clinicians specify a dictionary comprising all clinical concepts that are relevant to the desired phenotype (e.g. Alcohol Abuse).[19]

Our predictive models replicate the process as described by Liao et al.[12] We represent each note by the number of occurrences of each of the CUIs. Due to the fact that cTAKES detects negations, we count the occurrences of negated and non-negated CUIs separately. These features are then transformed to continuous features using the term frequency–inverse document frequency (TF-IDF). Compared to the original representation, or the bag-of-words of a note as described by Halpern et al.,[15] the TF-IDF of the features reflects the importance of a feature to a note. For an accurate comparison to approaches in literature, we train both a random forest (RF) and a logistic regression (LR) model with these features.

## Convolutional Neural Networks

Our proposed model is a convolutional neural network (CNN) for text classification, replicating the architecture proposed by Collobert et al. and Kim.[32,56] The idea behind convolutions in computer vision is to learn a transformation of adjacent pixels into a single value, similar to a filter.[57] In natural language processing, the model learns which combinations of subsequent words are associated with a given concept. An overview of our architecture is shown in Figure 1.

A major advantage of CNNs is that words in a text are first projected into distributed representations, often referred as word embeddings. Word embeddings have shown to improve performance on other tasks based on EHRs, for example NER.[58] Words that occur in similar contexts are trained to have similar word embeddings. Therefore, misspellings, synonyms and abbreviations of an original word learn similar embeddings, which lead to similar results. Consequently, a database of synonyms and common misspellings is not required.[19] Word embeddings can be pre-trained on a larger corpus of texts, which improves results of the NLP system and reduces the amount of data required to train a model.[59,60] We pre-train our embeddings with word2vec[61] on all discharge notes available in the MIMIC-III database.

The word embeddings of all words in the text to classify are concatenated and used as input to the convolutional layer. Convolutions detect a signal from a combination of adjacent inputs. We combine multiple convolutions of different lengths to evaluate phrases that are anywhere from two to five words long, as illustrated in Figure 1. The combination of many filters of varying length results in multiple outputs, which are then combined with max-pooling. More specifically, we use max-over-time-pooling to extract the most predictive value per filter.[56] The resulting prediction of the model utilizes a linear combination of these pooled features with a sigmoid function similar to a logistic regression.

*Figure 1: The architecture of our CNN model to perform the patient phenotyping. (A) Each word within a discharge note is looked up within a table of word embeddings and maps to its embedding. In our example, both instances of the word "and" will have the same embedding. (B) Convolutions of different widths are used to learn a filter that is applied to all embeddings of word sequences of the corresponding length. The convolution K2 with width 2 in our example of sequence length 11 will look at all 10 combinations of neighboring two words and output one value each. (C) The resulting multiple vectors are reduced to a single one using max-over-time pooling which will detect the highest value (the one with the most signaling power) for each of the different convolutions. (D) The final prediction ("Does the phenotype apply to the patient who the note belongs to?'") is made by computing a weighted combination of the pooled values and applying a sigmoid function, similar to a logistic regression. This figure is adapted with permission from Kim.[32]*

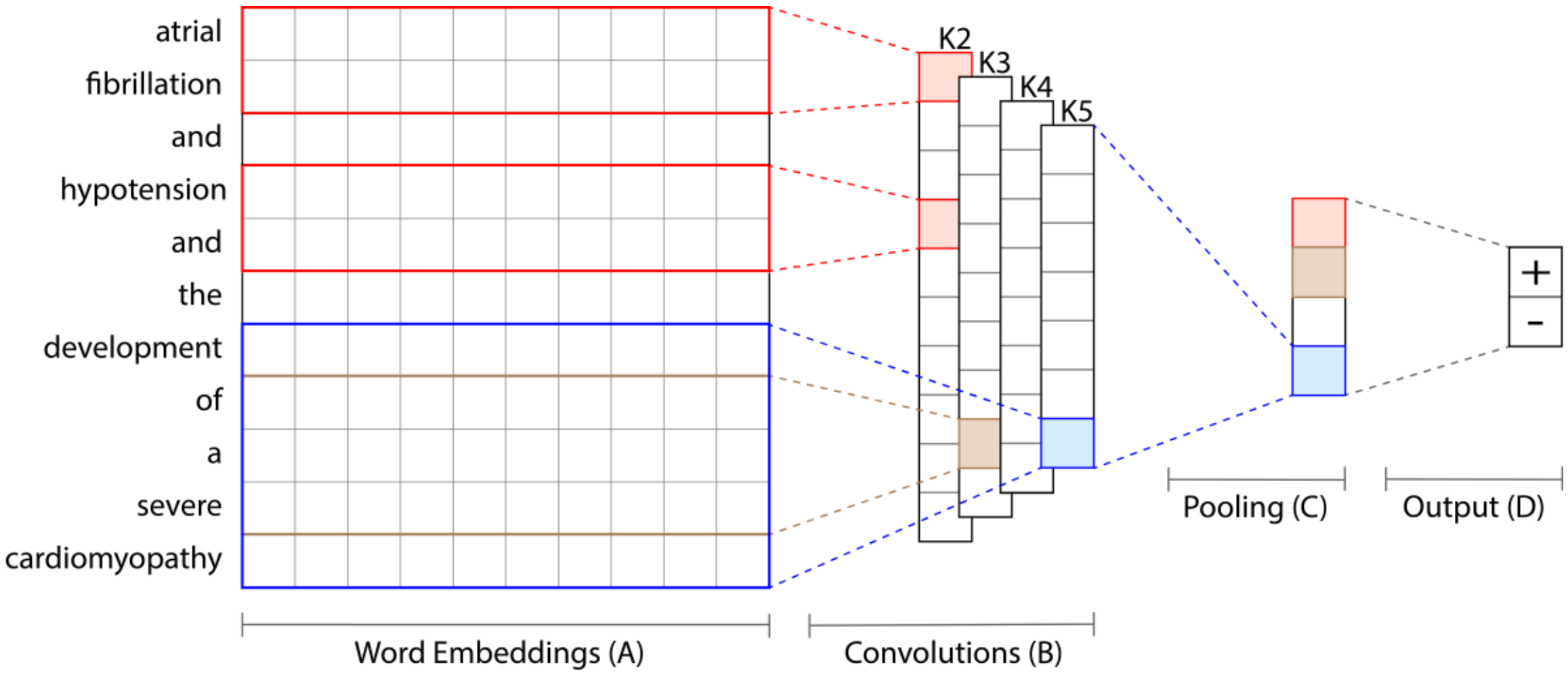

# DATA SET

All notes for this study are extracted from the MIMIC-III database. MIMIC-III contains de-identified clinical data of over 53,000 hospital admissions for adult patients to the intensive care units (ICU) at the Beth Israel Deaconess Medical Center from 2001 to 2012.[3] MIMIC-III includes several types of clinical notes, including discharge summaries (n = 52,746) and nursing notes (n=812,128).[62] In this study, we focus on the discharge summaries since they are the most informative for patient phenotyping.

We investigate phenotypes that may be associated with being a 'frequent flier' in the ICU (defined as >3 ICU visits within 365 days). As many as one third of readmissions have been suggested to be preventable; identifying modifiable risk factors is a crucial step to reducing them.[63] We extracted the first discharge summary from 415 ICU frequent fliers in MIMIC-III, as well as 313 randomly selected summaries from subsequent visits. We additionally selected 882 random summaries, yielding a total of 1,610 notes. The cTAKES output for these notes contains a total of 11,094 unique CUIs.

All 1,610 discharge summaries were annotated by clinicians for the 10 phenotypes shown in Table 1. Annotators for this study included two clinical researchers who have taken The Medical College Admission Test (MCAT®) (ETM, JW), two junior medical residents (JF, JTW), two senior medical residents (DWG, PDT), and a practicing intensive care medicine physician (LAC). The table shows the definition for each phenotype the annotating clinicians were instructed to look for, to improve inter-rater reliability. To ensure high-quality labels and minimize the risk of error, each note was labeled at least twice for each phenotype. In case the annotators were unsure, one of the senior clinicians (DWG or PDT) decided on the final label. The resulting number of occurrences of the phenotypes varies from 126 to 460 cases.

*Table 1: Definitions of phenotypes as well as the number of occurrences.*

| **Phenotype** | **#positive** | **Definition** |
|---|---|---|
| Adv. / Metastatic Cancer | 161 | Cancers with very high or imminent mortality (pancreas, esophagus, stomach, cholangiocarcinoma, brain); mention of distant or multi-organ metastasis, where palliative care would be considered (prognosis < 6 months). |
| Adv. Heart Disease | 275 | Any consideration for needing a heart transplant; Description of severe aortic stenosis (aortic valve area < 1.0cm^2), severe cardiomyopathy (LVEF <= 30%). Not sufficient to have a past medical history of congestive heart failure (CHF) or myocardial infarction (MI) with stent or coronary artery bypass graft (CABG) as these are too common. |
| Adv. Lung Disease | 167 | Severe chronic obstructive pulmonary disease (COPD) defined as Gold Stage III-IV or FEV1 < 50% of normal, or FEV1/FVC < 70%, or severe interstitial lung disease (ILD or IPF). |
| Chronic Neurological Dystrophies | 368 | Any chronic central nervous system (CNS) or spinal cord diseases, included/not limited to: Multiple sclerosis (MS), amyotrophic lateral sclerosis (ALS), myasthenia gravis, Parkinson's Disease, epilepsy, "previous history" of stroke/cerebrovascular accident (CVA) with residual deficits, and various neuromuscular diseases/dystrophies. |
| Chronic Pain | 321 | Any etiology of chronic pain, including fibromyalgia, requiring long-term opioid/narcotic analgesic medication to control. |
| Alcohol Abuse | 196 | Current/recent alcohol abuse history; still an active problem at time of admission (may or may not be the cause of it). |
| Substance Abuse | 155 | Include any intravenous drug abuse (IVDU), accidental overdose of psychoactive or narcotic medications (prescribed or not). Admitting to marijuana use in history not sufficient. |
| Obesity | 126 | Clinical obesity. BMI > 30. Previous history of or being considered for gastric bypass. Insufficient to have abdominal obesity mentioned in physical exam. |
| Psychiatric Disorders | 295 | All psychiatric disorders in DSM-5 classification, including schizophrenia, bipolar and anxiety disorders, other than depression. |
| Depression | 460 | Diagnosis of depression; prescription of anti-depressant medication; or any description of intentional drug overdose, suicide or self-harm attempts. |

## Performance Metrics

We evaluate the PPV, sensitivity, and F-score of all models. The F-score can be derived from a confusion matrix for the results on the test set. A confusion matrix contains four counts: true positive (TP), false positive (FP), true negative (TN), and false negative (FN). The PPV is the fraction of correct predictions out of all the samples that were predicted to be in a given category. The sensitivity, also known as recall, is the percentage of positive predictions in relation to all the predictions that should have been predicted as positive. The F-score is the harmonic mean of both PPV and sensitivity (more weight can be put on either of the two but we give equal weight to both, i.e. we use the F1-score).

$$PPV = \frac{TP}{TP + FN}$$

$$\text{Sensitivity } S = \frac{TP}{TP + FN}$$

$$\text{F1-Score } F = 2 * \frac{PPV * S}{PPV + S}$$

## Evaluation

For all of our models, we split the data into a training, validation and test set. 70% of the labeled data was used as the training set, 10% as validation set and 20% as test set. All reported numbers are obtained from testing on the same test set. The validation set is used to choose the hyperparameters of the models.

To achieve a fair comparison between the different types of models, we compare different approaches for each. We compare the performance of all models to two simple baselines based on n-gram models to check that a more complicated model such as a CNN is actually necessary or whether simple co-occurrences of words can pick up the signals. Therefore, we report numbers on the eight models and baselines shown in Table 2.

*Table 2: Descriptions of our different models and baselines.*

| **Model Name** | **Description of the Model** |
|---|---|
| CNN | Our proposed convolutional neural network architecture |
| 2-gram LR | Baseline that uses bigrams of a text as input to a logistic regression |
| 3-gram LR | Baseline that uses trigrams of a text as input to a logistic regression |
| cTAKES RF | Random forest that uses the full output from cTAKES |
| cTAKES LR | Logistic regression that uses the full output from cTAKES |
| Filter RF | Random forest that uses clinician-filtered output from cTAKES |
| Filter LR | Logistic regression that uses clinician-filtered output from cTAKES |

For the CNN model, we used 100 filters for each of the widths 2, 3, 4, and 5. To prevent over-fitting, we set the dropout probability to 0.5 and used L2-normalization to normalize word embeddings to have a max norm of 3.[64] The model was trained using adadelta with an initial learning rate of 1 for 20 epochs.[65] The CNN model was implemented using Lua and the Torch7 framework.[66] All baseline models were implemented using Python with the scikit-learn library.[67]

## Interpretability

We compare the interpretability of the approaches by assessing which phrases are the most salient for a positive prediction on a global model-wide scale. We evaluate the Filter LR model, because its learned parameters that correspond to each CUI are a direct indication of how salient it is and irrelevant CUIs are already removed by clinicians, making sure that all CUIs are relevant.[21] For the CNN, we compute a modified saliency of all phrases. The saliency in neural networks is defined as the norm of the gradient of the loss function with respect to an input.[68] Alternative methods search the local space around an input[27], or compute a layer-wise backpropagation.[69–72] An input in our case is a single word embedding; to evaluate the whole phrase we calculate the norm of the convolutional layer for positive predictions instead. This approximates how much a phrase contributed to a prediction and works well in our case. To obtain the most important phrases globally, we classify and evaluate all documents in the test set and store the most indicative phrases. To assess the saliency on a local document level, we can extract the most indicative phrases that exist in a given document using the same methodologies.

# RESULTS

We show an overview of the F1-scores for different models and phenotypes in Figure 2. For every phenotype, the CNN outperforms all other approaches. For some of the phenotypes such as *Obesity* and *Schizophrenia*, the CNN outperforms the other models by a large margin. The filtered models, which require much more effort from clinicians, only have a minimal improvement over the noisy input of all identified cTAKES concepts.

In the detailed results, shown in Table 3, we observe that the CNN outperforms the baselines on all of the sensitivity values and half of the PPV's. In some cases, the simple n-gram baselines achieve a very high PPV with a very low sensitivity. That means that these models could be efficiently used to detect patients if it does not matter that the model overlooks most of the positives, for example to detect a small at-risk population for interventions.[73]

*Figure 2: Comparison of achieved F1-scores across all tested phenotypes. Our models are the 3 models on the left of each phenotype, shown in blue. The 4 cTAKES-based models are on the right, in red. The CNN achieves the highest scores across all phenotypes.*

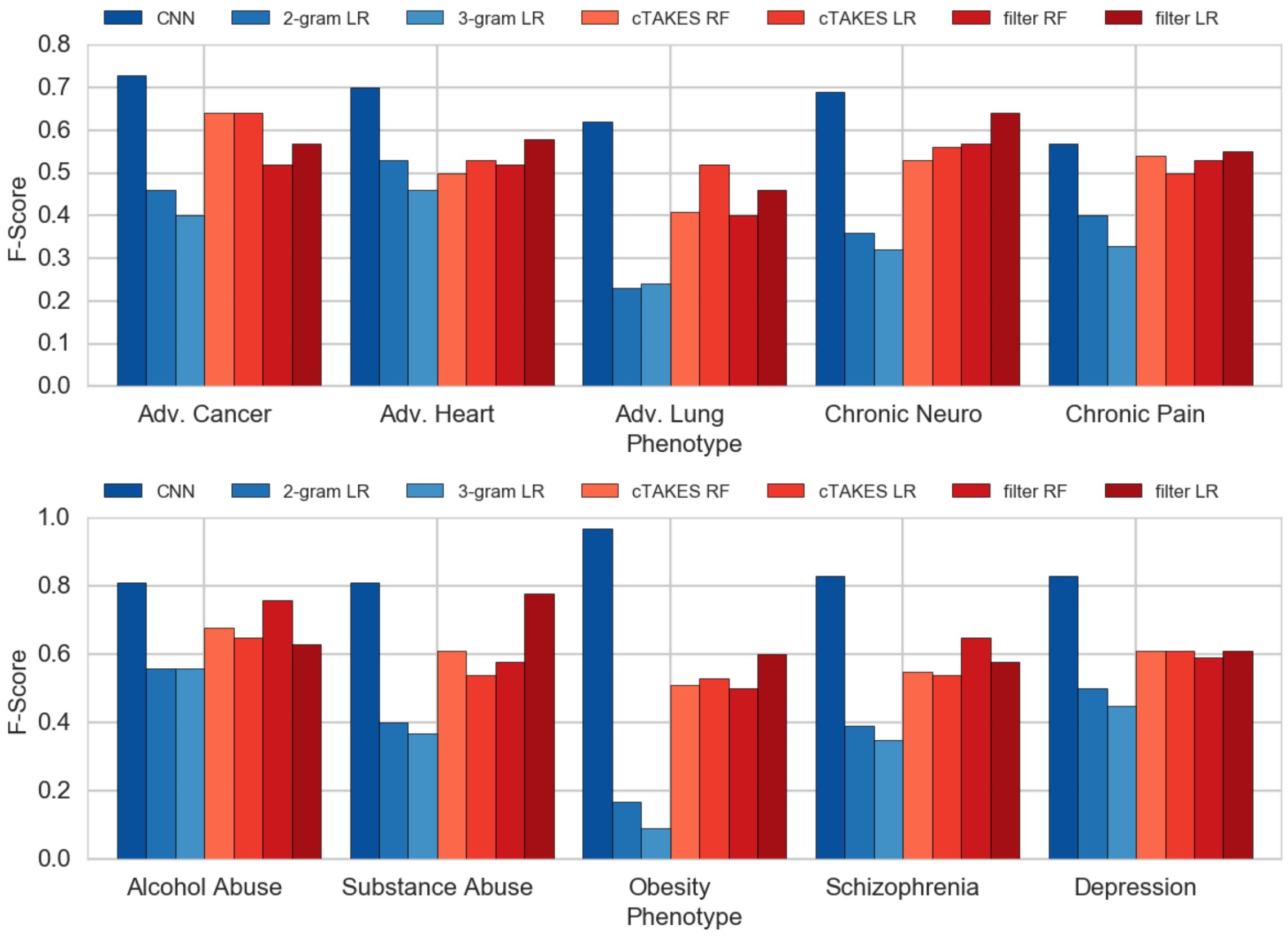

We show the most salient phrases according to the CNN and the filtered cTAKES LR models for *Advanced Heart Disease* in Table 4, and for *Alcohol Abuse* in Table 5. Both tables contain many phrases mentioned in the definition shown in Table 1, such as "*Cardiomyopathy*". We also observe mentions of "*CHF*" and "*CABG*" in Table 4 for both models, which are common medical conditions associated with advanced heart disease, but are not sufficient requirements to be diagnosed or annotated as such in the annotation scheme. The model still learned to associate those phrases with advanced heart disease, since those phrases also occur in many notes from patients that were labeled positive for advanced heart failure. We argue that overall, there is no loss in interpretability when looking at the learned phrases. Moreover, while the CUIs extracted by cTAKES can be very generic, such as "*Atrium, Heart*" or "*Heart*", the salient CNN phrases are more specific.

The phrases in Table 5 illustrate how the CNN can detect mentions of the condition in many forms. Without any human input, the CNN learned that *EtOH* and *alcohol* are used synonymously and thus detects phrases containing either of them, which leads to a higher sensitivity. The filtered cTAKES LR model surprisingly ranks *victim of abuse* higher than the direct mention of *alcohol abuse* in a note, and finds it very indicative if an ethanol measurement was taken.

*Table 3: The results (PPV, Sensitivity, and F1-score) across all phenotypes for all models. cT stands for cTAKES.*

| **Phenotype** | | **CNN** | **2-gram** | **3-gram** | **cT RF** | **cT LR** | **Filter RF** | **Filter LR** |
|---|---|---|---|---|---|---|---|---|
| Adv. Cancer | *PPV* | 90 | 91 | **100** | 94 | 94 | 68 | 78 |
| | *S* | **61** | 31 | 25 | 48 | 48 | 42 | 45 |
| | *F* | **73** | 46 | 40 | 64 | 64 | 52 | 57 |
| Adv. Heart Disease | *PPV* | 73 | 69 | 71 | 56 | 65 | 58 | **74** |
| | *S* | **68** | 43 | 34 | 46 | 44 | 47 | 47 |
| | *F* | **70** | 53 | 46 | 50 | 53 | 52 | 58 |
| Adv. Lung Disease | *PPV* | **67** | 57 | **67** | 36 | **67** | 38 | 46 |
| | *S* | **57** | 14 | 14 | 46 | 43 | 43 | 46 |
| | *F* | **62** | 23 | 24 | 41 | 52 | 40 | 46 |
| Chronic Neurological | *PPV* | **81** | 56 | 55 | 58 | 66 | 70 | 87 |
| | *S* | **61** | 27 | 23 | 49 | 49 | 49 | 51 |
| | *F* | **69** | 36 | 32 | 53 | 56 | 57 | 64 |
| Chronic Pain | *PPV* | **78** | 49 | 44 | 61 | 53 | 62 | 68 |
| | *S* | 45 | 33 | 26 | **48** | **48** | 46 | 46 |
| | *F* | **57** | 40 | 33 | 54 | 50 | 53 | 55 |
| Alcohol Abuse | *PPV* | 85 | **100** | **100** | 94 | 76 | **100** | **100** |
| | *S* | **79** | 39 | 39 | 54 | 57 | 61 | 46 |
| | *F* | **81** | 56 | 56 | 68 | 65 | 76 | 63 |
| Substance Abuse | *PPV* | 83 | 80 | 88 | 79 | 64 | 87 | **95** |
| | *S* | **80** | 27 | 23 | 50 | 47 | 43 | 67 |
| | *F* | **81** | 40 | 37 | 61 | 54 | 58 | 78 |
| Obesity | *PPV* | **100** | 50 | 50 | 60 | 80 | 67 | 90 |
| | *S* | **95** | 10 | 5 | 45 | 40 | 40 | 45 |
| | *F* | **97** | 17 | 9 | 51 | 53 | 50 | 60 |
| Psychiatric Disorders | *PPV* | 87 | 61 | 67 | 62 | 62 | **88** | 79 |
| | *S* | **80** | 29 | 24 | 49 | 47 | 51 | 46 |
| | *F* | **83** | 39 | 35 | 55 | 54 | 65 | 58 |
| Depression | *PPV* | **91** | 67 | 67 | 82 | 77 | 74 | 82 |
| | *S* | **76** | 40 | 34 | 49 | 50 | 49 | 49 |
| | *F* | **83** | 50 | 45 | 61 | 61 | 59 | 61 |

*Table 4: The most salient phrases for Advanced Heart Failure. The salient cTAKES CUIs are extracted from the filtered LR model. Duplicate phrases are removed.*

| Most relevant cTAKES CUIs | Most salient phrases detected by the CNN |
|---|---|
| *Magnesium* | *Wall Hypokinesis* |
| *Cardiomyopathy* | *Port pacer* |
| *Hypokinesia* | *Ventricular hypokinesis* |
| *Heart Failure* | *p AVR* |
| *Acetylsalicylic Acid* | *post ICD* |
| *Atrium, Heart* | *status post ICD* |
| *Coronary Disease* | *EF 20 30* |
| *Atrial Fibrillation* | *bifurcation aneurysm clipping* |
| *Coronary Artery* | *CHF with EF* |
| *Disease* | *cardiomyopathy , EF 15* |
| *Aortocoronary Bypasses* | *( EF 20 30* |
| *Fibrillation* | *coronary artery bypass graft* |
| *Heart* | *respiratory viral infection by DFA* |
| *Catheterization* | *severe global free wall hypokinesis* |
| *Chest* | *Class II , EF 20* |
| *Artery* | *lateral CHF with EF 30* |
| *CAT Scans, X-Ray* | *anterior and atypical hypokinesis akinesis* |
| *Hypertension* | *severe global left ventricular hypokinesis* |
| *Creatinine Measurement* | *'s cardiomyopathy , EF 15* |

*Table 5: The most salient phrases for Alcohol Abuse. The salient cTAKES CUIs are extracted from the filtered LR model. Duplicate phrases are removed.*

| Most relevant cTAKES CUIs | Most salient phrases detected by the CNN |
|---|---|
| *Victim of abuse* | *Consciousness Alert* |
| *Ethanol Measurement* | *Alcohol Abuse* |
| *Alcohol Abuse* | *EtOH abuse* |
| *Thiamine* | *Alcoholic Dilated* |
| *Social and personal history* | *ETOH cirrhosis* |
| *Family history* | *heavy alcohol abuse* |
| *Hypertension* | *evening Alcohol abuse* |
| *Injuries risk* | *Drug Reactions Attending* |
| *Pain* | *alcohol withdrawal compartment syndrome* |
| *Sodium* | *EtOH abuse with multiple* |
| *Potassium Measurement* | *liver secondary to alcohol abuse* |
| *Plasma Glucose Measurement* | *abuse crack cocaine, EtOH* |

# DISCUSSION

CNNs are a novel and flexible approach to patient phenotyping using clinical notes. Our results show that deep learning outperforms all other methods in terms of F1-score and sensitivity while achieving a comparable or better PPV. However, we notice that even consistent annotation schemes lead to varying results between phenotypes. This makes it especially difficult to compare our results to other reported metrics in the literature, a problem that is amplified by the sparsity of available studies using only unstructured data.

The major advantage of rule-based models that are specifically tailored to a given problem is their interpretability. Clinicians dictate the phrases that are considered as input to a classifier and have therefore full control over the model. Since bias in data collection and analysis is at times unavoidable, models are required to be interpretable in order for clinicians to be able to detect such biases. One such example of bias was in mortality prediction among patients with pneumonia where asthma was found to increase survival probability.[25] It turned out that there was an institutional practice to admit all patients with pneumonia and a history of asthma to the ICU regardless of disease severity, so that a history of asthma was strongly correlated with a lower illness severity.

We demonstrated that CNNs can be interpreted in the same way as rule-based models by computing the saliency of inputs. This leads to a similar level of interpretability. One disadvantage of our approach is the requirement for more phrases for consideration. Lists of salient phrases will naturally comprise more items, making it more difficult to investigate which phrases exactly lead to a prediction. However, each phrase comes with a saliency coefficient, which allows compensating for the length of the list of salient phrases.

Another point of comparison is the complexity of the annotation task involved for clinicians, who may not be familiar with NLP data set creation methodologies. Both the CNN and rule-based approaches require the construction of an annotated data set, a process that can span from several months to years, especially if the labels cannot be inferred from the structured data itself. Since a CNN learns about phrases in notes that are associated with the presence of a concept, the clinicians can simply indicate the presence or absence of a concept of interest while guided by clinically driven criteria. As such, our proposed approach allows annotations with broader diagnostic criteria instead of limiting the annotation rules to specific pre-defined phrases. This annotation approach is more suitable for modeling concepts that require interpretation of complex contextual or clinical reasoning patterns. While a CNN learns the rules that lead to a positive label, rule-based approaches require clinicians to define every phrase that is associated with a concept. Due to the heterogeneity of text, clinicians might not be able to think of all possible phrases in advance. They also have to consider how to handle negated phrases correctly. Finally, for some clinically important phenotypes such as “Non-Adherence”, it is impossible to construct an exhaustive list of phrases associated with it.

There are some limitations for the CNN. Because CNNs learns the phrases associated with positively annotated notes, the algorithm’s generalizability is even more dependent on the initial note selection criteria for its training data. Additionally, our approach still requires an annotated data set. Therefore, cTAKES-based models may still be preferable for applications where a lower sensitivity is acceptable.

However, the advantage of the CNN, and the annotation strategy that it enabled, lies in allowing rapid development of phenotyping capabilities for multiple complex concepts simultaneously from only unstructured data. Our annotation strategy takes approximately the same amount of time to annotate any number of concepts once a clinician is reading a note. While rule-based systems require a separate algorithm for each annotation, the same CNN can be trained for all of the annotated phenotypes at the same time. This offers an opportunity to dramatically accelerate the development of high-quality corpora of annotated data as well as scalable phenotyping algorithms. Such capabilities are important for identifying complex clinical concepts in unstructured clinical text that are poorly captured in the structured data. For example, being able to identify patients who are readmitted to hospital due to poor management of problems, such as drug abuse, psychiatric disorders and other chronic diseases, which are often poorly coded, will have high clinical impact.

As we mentioned before, the goal with our data is to understand phenotypes that are indicative of patients having repeated ICU admissions. Due to the multiple phenotypes that are hypothesized to be associated with it, we require a deep learning algorithm that supports this rapid phenotyping. Additionally, we anticipate validation of this approach in other types of clinical notes such as social work assessment to identify patients at risk. Lastly, the CNN creates the opportunity to develop a model that can use phrase saliency to highlight notes and tag patients to support chart review. We are planning future work to explore whether the identification of salient phrases can be used to support chart abstraction and whether models using these phrases represent what clinicians find salient in a medical note.

# CONCLUSION

We have presented an alternative approach to patient phenotyping using NLP based on deep learning. Our model significantly improves the accuracy of phenotyping while decreasing the annotation complexity required of clinical domain experts. Our approach can be employed to augment structured data in the EHR for a variety of phenotyping tasks. We address concerns about the interpretability of deep learning by proposing a method to identify phrases associated with different phenotypes.

## ACKNOWLEDGEMENTS
We thank Barbara J. Grosz for helpful discussions.

## FUNDING INFORMATION
Franck Dernoncourt is supported by a grant from Philips Research. Leo Anthony Celi is supported by the R01 grant EB017205-01A1 from the National Institute of Bioimaging and Biomedical Engineering. The content is solely the responsibility of the authors and does not necessarily represent the official views of Philips Research or the National Institute of Bioimaging and Biomedical Engineering.

## COMPETING INTERESTS
We have no competing interests to declare.